\documentclass{llncs}
\usepackage{graphicx}
\usepackage{multirow}
\usepackage{natbib}

\usepackage[utf8]{inputenc}
\usepackage[colorlinks,citecolor=blue,linkcolor=blue]{hyperref}
\usepackage{tikz}
\usepackage{comment}
\usepackage{amsmath,amssymb} % define this before the line numbering.
\usepackage{color}

\usepackage[accsupp]{axessibility}  % Improves PDF readability for those with disabilities.

\usepackage[left=0.9in,right=0.9in,top=0.9in,bottom=0.9in]{geometry}

\usepackage{fancyhdr}

\usepackage{blindtext}
\begin{document}
% \linenumbers

% \title{Fine-tuning Agricultural Models on Agricultural Data Increases Efficiency and Performance\\[1em]Data Efficient Domain-Specific Method Adaptations for Agricultural Deep Learning Models\\[1em]Assessing Data Efficient Deep Learning Methods on Standardized Agricultural Datasets\\[1em]Standardizing and Centralizing Agricultural Datasets and Developing Efficient Training Methods for Agricultural Deep Learning Models\\[1em]Adapting Methods for Training Deep Learning Models and Establishing Benchmarks on Standardized Agricultural Datasets
% } % Replace with your title

\makeatletter
\def\blfootnote{\gdef\@thefnmark{}\@footnotetext}
\makeatother

\title{Standardizing and Centralizing Datasets to Enable Efficient Training of Agricultural Deep Learning Models}

% \begin{comment}
\author{Amogh Joshi\inst{1,2,3} \and Dario Guevara\inst{1,2,3} \and Mason Earles\inst{1,2,3}}
\institute{Department of Viticulture and Enology, University of California, Davis \and Department of Biological and Agricultural Engineering, University of California, Davis \and AI Institute for Next-Generation Food Systems (AIFS)\\\email{\{amnjoshi,dguevara,jmearles\}@ucdavis.edu}}
% \end{comment}
%******************

\begin{comment}
\titlerunning{AgML Public Datasets and Benchmarking}
\author{First Author\inst{1}\orcidID{0000-1111-2222-3333} \and
Second Author\inst{2,3}\orcidID{1111-2222-3333-4444} \and
Third Author\inst{3}\orcidID{2222--3333-4444-5555}}
%
\authorrunning{F. Author et al.}
\institute{Princeton University, Princeton NJ 08544, USA \and
Springer Heidelberg, Tiergartenstr. 17, 69121 Heidelberg, Germany
\email{lncs@springer.com}\\
\url{http://www.springer.com/gp/computer-science/lncs} \and
ABC Institute, Rupert-Karls-University Heidelberg, Heidelberg, Germany\\
\email{\{abc,lncs\}@uni-heidelberg.de}}
\end{comment}
%******************
\maketitle

\begin{abstract}

In recent years, deep learning models have become the standard for agricultural computer vision. Such models are typically fine-tuned to agricultural tasks using model weights that were originally fit to more general, non-agricultural datasets. This lack of agriculture-specific fine-tuning potentially increases training time and resource use, and decreases model performance, leading an overall decrease in data efficiency. To overcome this limitation, we collect a wide range of existing public datasets for three distinct tasks, standardize them, and construct standard training and evaluation pipelines, providing us with a set of benchmarks and pretrained models. We then conduct a number of experiments using methods which are commonly used in deep learning tasks, but unexplored in their domain-specific applications for agriculture. Our experiments guide us in developing a number of approaches to improve data efficiency when training agricultural deep learning models, without large-scale modifications to existing pipelines. Our results demonstrate that even slight training modifications, such as using agricultural pretrained model weights, or adopting specific spatial augmentations into data processing pipelines, can significantly boost model performance and result in shorter convergence time, saving training resources. Furthermore, we find that even models trained on low-quality annotations can produce comparable levels of performance to their high-quality equivalents, suggesting that datasets with poor annotations can still be used for training, expanding the pool of currently available datasets. Our methods are broadly applicable throughout agricultural deep learning, and present high potential for significant data efficiency improvements.

% \keywords{Agricultural Deep Learning, Data Efficiency, Image Augmentation, Transfer Learning, Model Benchmarking}
\end{abstract}

\section{Introduction}

Deep learning models have become standard for modern computer-vision-based agricultural tasks. Examples of common standard tasks now largely automated by deep learning include fruit detection~\citep{fruit_detection_worldwide,deep_orchard_detection}, crop and weed segmentation~\citep{robust_crop_weed_segmentation,real_time_semantic_segmentation_crop_weed}, and plant disease classification~\citep{deep_learning_plant_disease_detection,efficientnet_disease_classification}. Certain deep learning models have even been used for tasks beyond standard single-RGB-image predictions, involving hyperspectral and thermal imagery for various analyses~\citep{visible_thermal_imagery,hyperspectral_deep_learning_review} or tasks involving multiple scales of data, such as spatiotemporal crop yield prediction~\citep{crop_yield_spatiotemporal}. Existing approaches to such tasks often involve direct applications of state-of-the-art models developed for other tasks in the field of deep learning, which have been thoroughly evaluated by works such as~\cite{survey_agricultural_models_detection}.

\begin{figure}[!h]
    \centering
    \includegraphics[width=0.9\columnwidth]{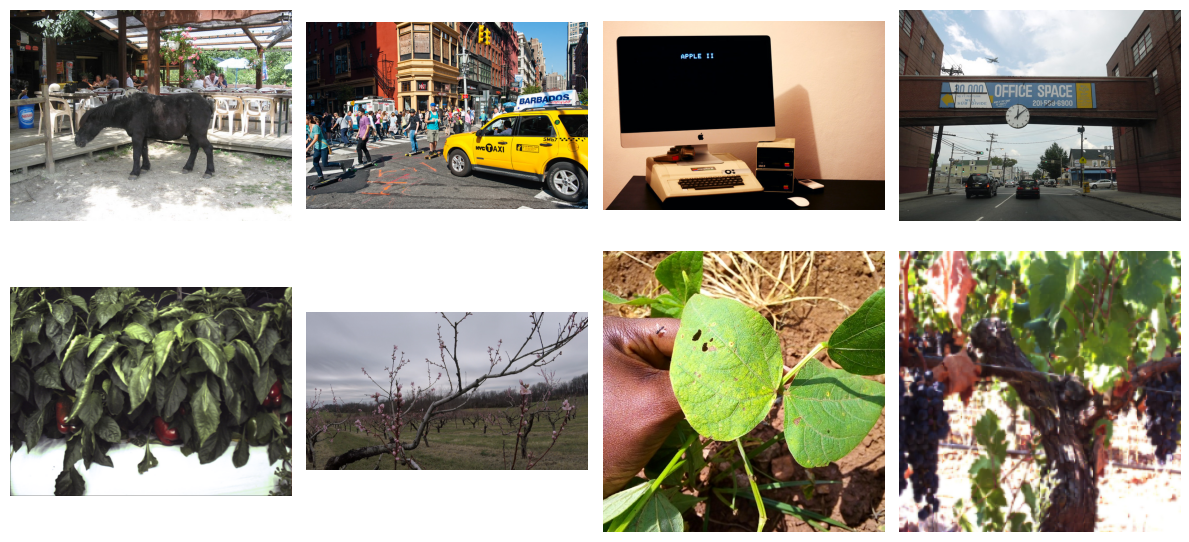}
    \caption{Sample of images from the COCO dataset (row 1) and from assorted agricultural datasets (row 2), showcasing the contrast between the general environment of COCO imagery and the specific environment of agricultural imagery.}
    \label{fig:coco_ag_example}
\end{figure}

A largely prevalent problem in the agricultural domain is a significant deficiency of task-specific data. Transfer learning~\citep{transfer_learning} is an approach undertaken when training large deep learning models which attempts to offset this data deficiency, therein transferring knowledge from a source task to the new, reduced size target dataset. In practice, this generally consists of replacing random weight initialization for model parameters with existing pretrained weights from a prior task, converting a complete training task into a finetuning task using existing pretrained models -- for instance, many image classification and object detection tasks start off with pretrained weights from the ImageNet~\citep{imagenet} and COCO~\citep{coco} datasets, respectively. However, these datasets are mostly generalized to common objects in environments ranging from inside a home to the streets of a city. Such environments consist of a broad range of objects and conditions, allowing models to locate specific objects with relatively greater ease. In contrast, agricultural environments are highly domain-specific, and these existing pretrained model approaches may not offer as significant of a knowledge transfer for agricultural domains. An example of the contrast between the aforementioned conditions is shown in Figure~\ref{fig:coco_ag_example}. This, in turn, suggests the potential need for alternative pretrained model approaches in order to improve data efficiency when training agricultural deep learning models. Previously,~\cite{crop_type_mapping_transfer}, ~\cite{korean_knowledge_transfer_greenhouse}, and~\cite{rice_plant_transfer_learing} applied transfer learning to agricultural image classification tasks, demonstrating improved performance using pretrained weights on the ImageNet dataset, citing its large magnitude of data and certain specific classes relevant to agriculture. However, little to no prior work has been conducted assessing the viability of transfer learning in other, more relevant agricultural tasks, like semantic segmentation and object detection. Furthermore, there does not exist a single centralized repository for agriculture-specific datasets, preventing a large-scale ImageNet-style dataset for agriculture from being developed. While work done by~\cite{cropdeep_paper} and~\cite{agrinet_paper} have proposed new, larger-scale datasets for agriculture, increasing progress towards such a goal, such datasets are usually specific to certain conditions and designed for image classification, leaving other tasks without as much data. As a result, the potential for transfer learning as a viable improvement for data efficiency, at least in agriculture, is diminished.

Another common approach in deep learning which attempts to offset a deficiency of data is the use of data augmentations. For computer vision tasks, this refers to the transformation of an input image along with its corresponding annotations, with the stated intent of increasing diversity of data. Standard augmentations include rotation or flipping of an input image, or adjustment of visual parameters like saturation or brightness. The usefulness of augmentations have been studied in a general case in works such as~\cite{image_augmentation_deep_learning_survey}; however these applications have largely been focused either on general image classification tasks or for a specific domain which is not similar with agriculture. Agricultural environments are largely different than general environments and are often more complex, involving less quantities of obvious features and more irregular and inconsistent shapes~\citep{ag_rs_multispectral_difficult_environment}. Certain studies of augmentations have been conducted in works such as~\cite{hsv_learning_tomato}, where custom colorspaces and vegetation indexes were used when processing input data in order to boost performance on agricultural datasets, but there has not been done any large-scale analysis of augmentation effectiveness for agricultural environments.  In fact, a potential benefit of the domain constraints of agricultural data is the prospect for assessing features specific to these environments. This can be used to both develop data processing pipelines specific to agriculture, using such approaches, and adapt existing model architectures to agricultural tasks using methods like transfer learning. Standing in the way of this development is a lack of a large-scale centralized database for agricultural data.

Bringing together datasets from previous studies, we present a novel set of centralized and standardized public agricultural datasets and benchmarks using state-of-the-art deep learning models. This collection of datasets is composed of real data collected from agricultural datasets across three different tasks: image classification, semantic segmentation, and object detection. For each task, we develop a standard data format and training pipeline, which we use to generate benchmarks and pretrained models for each dataset.These pipelines and models enable us to conduct a set of experiments in which we adapt existing methods for improving data efficiency, such as transfer learning and image augmentation, as described above, for agricultural environments. We first conduct an assessment of agricultural models for transfer learning, by (1) assessing data efficiency and performance when using agricultural pretrained weights for fruit detection models. We apply this study further by (2) assessing data efficiency and performance when using pretrained weights for backbones of agricultural semantic segmentation models for fruit and plant segmentation. These two experiments provide us with a quantifiable way to improve data efficiency for agricultural models. Following our analysis of methods to improve models, we continue our study by evaluating methods to improve data, by (3) assessing the effectiveness of standard spatial and visual image augmentations for improving model performance on fruit detection and fruit segmentation tasks. Finally, we note that while our methods suggest methods to improve data, there exists a pool of low-quality data which may potentially be disregarded, we (4) assess the effect of annotation quality, primarily accuracy and quantity of bounding boxes, on fruit detection tasks. Through these experiments, we find a strong potential for improving data efficiency for agricultural deep learning models without significant modifications to existing pipelines, with potential implications for a broad range of tasks. Furthermore, we open-source our standardized datasets and pretrained models through our framework AgML\footnote{\url{https://github.com/Project-AgML/AgML}}, to facilitate the adoption of our described methods.

% \begin{itemize}
%     \item Agricultural machine learning and computer vision; datasets, benchmarks, models, evaluation
%     \item Transfer learning and model pre-training; Link broader transfer learning challenges to more specific ag domain opportunities
%     \item Fruit detection, and links to transfer learning; this should set up the context for the experiments that you've been doing
%     \item End here with a short paragraph setting up our approach
% \end{itemize}

\section{Methods and Experiments}

To provide a baseline for our experiments, we have collected a set of agricultural datasets and developed a novel set of standard benchmarks and pretrained models on these datasets. These datasets have been standardized by task and are included in the dataset catalogue in the open-source agricultural machine learning framework called AgML. We discuss our methodology in developing a standard task-based data processing pipeline for training agricultural machine learning models, and our models' results on these datasets. These pipelines are designed to be adaptable for the experiments conducted later in this research, and serve as our primary tool for them.

\subsection{Collecting Datasets}

\begin{figure}[h]
    \centering
    \includegraphics[width=0.7\columnwidth]{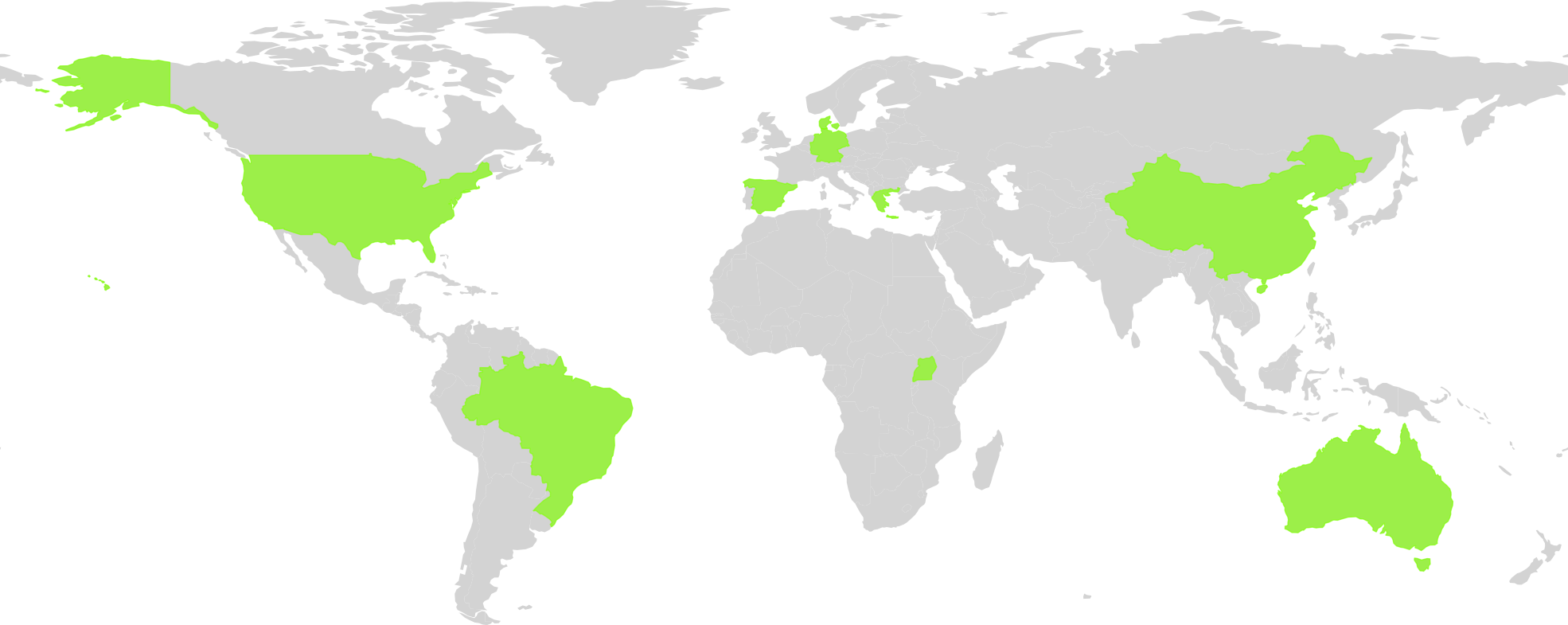}
    \caption{Countries from which agricultural data in the AgML public dataset collection was sourced.}
    \label{fig:data_countries}
\end{figure}

For the purposes of this work, we restrict datasets to three tasks: image classification, semantic segmentation, and object detection. Furthermore, we focus on datasets with only a single 3-channel RGB image input, in order to facilitate the creation of benchmarks for a variety of datasets using standard deep learning models without any architectural modifications, allowing easier accessibility for a broader range of future applications. We compile datasets from a number of sources, with the majority coming from~\cite{survey_of_public_datasets}. We provide a full listing of each of the datasets used in this dataset as well as key details and references in Table~\ref{table:dataset_listing_with_benchmarks}, which also contains benchmark performance on metrics as detailed later in Section~\ref{sec:models_and_methods}.

\setlength{\tabcolsep}{2pt}
\begin{table}[!t]
% \small
\begin{center}
\caption{A listing of publicly available agricultural datasets used as part of this research (as described in \textit{Dataset Name}), along with their number of images (\textit{Images}), agricultural task (\textit{Ag. Task}), and corresponding benchmark performance for the models described above on a testing set (\textit{Benchmark})}
\label{table:dataset_listing_with_benchmarks}
\begin{tabular}{lrrcr}
\hline\noalign{\smallskip}
Dataset Name & Images & Classes & Ag. Task & Benchmark \\
\hline
\noalign{\smallskip}
\hline
\noalign{\smallskip}
{\bf Image Classification} & & & & \texttt{Accuracy}\\\hline
\noalign{\smallskip}

\texttt{bean\textunderscore disease\textunderscore uganda} & 1295 & 3 & \textit{B} & \textbf{96.90\%} \\
\texttt{plant\textunderscore seedlings\textunderscore aarhus}~\citep{plant_seedlings_aarhus} & 5539 & 12 & \textit{A} & \textbf{94.39\%}\\
\texttt{soybean\textunderscore weed\textunderscore uav\textunderscore brazil}~\citep{soybean_weed_uav_brazil} & 15336 & 4 & \textit{A} & \textbf{100.0\%}\\
\texttt{sugarcane\textunderscore damage\textunderscore usa}~\citep{sugarcane_damage_usa} & 153 & 6 & \textit{B} & \textbf{100.0\%}\\
\texttt{crop\textunderscore weeds\textunderscore greece}~\citep{crop_weeds_greece} & 508 & 4 & \textit{A} & \textbf{100.0\%}\\
\texttt{rangeland\textunderscore weeds\textunderscore australia}~\citep{rangeland_weeds_australia} & 17509 & 10 & \textit{A} & \textbf{97.94\%}\\
\texttt{leaf\textunderscore counting\textunderscore denmark}~\citep{leaf_counting_denmark} & 9372 & 9 & \textit{C} & \textbf{88.90\%} \\
\texttt{plant\textunderscore village\textunderscore classification}~\citep{plant_village_classification} & 55448 & 39 & \textit{B} & \textbf{98.91\%}\\
\texttt{plant\textunderscore doc\textunderscore classification}~\citep{plant_doc_classification} & 2336 & 28 & \textit{B} & \textbf{89.27\%}\\

\hline
\noalign{\smallskip}
{\bf Semantic Segmentation} & &  & & \texttt{mIoU}\\\hline
\noalign{\smallskip}

\texttt{carrot\textunderscore weeds\textunderscore germany}~\citep{carrot_weeds_germany} & 60 & 2 & \textit{D} & \textbf{52.18\%}\\
\texttt{sugarbeet\textunderscore weed\textunderscore segmentation}~\citep{sugarbeet_weed_segmentation} & 125 & 2 & \textit{D} & \textbf{53.59\%}\\
\texttt{apple\textunderscore flower\textunderscore segmentation}~\citep{apple_flower_segmentation} & 148 & 
1 &\textit{E} & \textbf{68.38\%}\\
\texttt{apple\textunderscore segmentation\textunderscore minnesota}~\citep{apple_segmentation_minnesota} & 670 & 1 &\textit{F} & \textbf{79.08\%}\\
\texttt{rice\textunderscore seedling\textunderscore segmentation}~\citep{rice_seedling_segmentation} & 224 & 2 &\textit{D} & \textbf{52.20\%}\\
\texttt{peachpear\textunderscore flower\textunderscore segmentation}~\citep{apple_flower_segmentation} & 42 & 1 & \textit{E} & \textbf{72.58\%}\\
\texttt{red\textunderscore grapes\textunderscore and\textunderscore leaves\textunderscore segmentation}~\citep{grapes_leaves_segmentation} & 258 & 2 & \textit{F} & \textbf{49.18\%}\\
\texttt{white\textunderscore grapes\textunderscore and\textunderscore leaves\textunderscore segmentation}~\citep{grapes_leaves_segmentation} & 273 & 2 & \textit{F} & \textbf{51.93\%}\\

\hline
\noalign{\smallskip}
{\bf Object Detection} & & & & \texttt{mAP@0.5}\\\hline
\noalign{\smallskip}

\texttt{fruit\textunderscore detection\textunderscore worldwide}~\citep{fruit_detection_worldwide} & 565 & 7 & \textit{G} & \textbf{70.35\%}\\
\texttt{apple\textunderscore detection\textunderscore usa}~\citep{apple_detection_usa} & 2290 & 1 & \textit{G} & \textbf{94.16\%}\\
\texttt{apple\textunderscore detection\textunderscore spain}~\citep{apple_detection_spain} & 967 & 1 & \textit{G} & \textbf{86.65\%}\\
\texttt{apple\textunderscore detection\textunderscore drone\textunderscore brazil}~\citep{apple_detection_drone_brazil} & 689 & 1 & \textit{G} & \textbf{79.62\%}\\
\texttt{mango\textunderscore detection\textunderscore australia}~\citep{mango_detection_australia} & 1242 & 1 & \textit{G} & \textbf{95.32\%}\\
\texttt{grape\textunderscore detection\textunderscore californiaday}~\citep{PAIBL} & 126 & 1 & \textit{G} & \textbf{69.01\%}\\
\texttt{grape\textunderscore detection\textunderscore californianight}~\citep{PAIBL}& 150 & 1 & \textit{G} & \textbf{63.99\%}\\

\hline
\noalign{\smallskip}
\hline
\noalign{\smallskip}
\multicolumn{5}{r}{\textbf{Task Legend} $\qquad\qquad\qquad\qquad\qquad\qquad\;\;\;\;\;\;^{A, B} \textrm{\{Weed, Disease/Damage\} Classification}$, $^{C}\textrm{Leaf Counting}$} \\[0.05em]
\multicolumn{5}{r}{$^{D, E, F}\textrm{\{Weed, Flower, Fruit\} Segmentation}$, $^G\textrm{Fruit Detection}$} \\
\hline
\end{tabular}
\end{center}
\end{table}
\setlength{\tabcolsep}{1.4pt}

\begin{figure}[h]
    \centering
    \includegraphics[width=0.95\columnwidth]{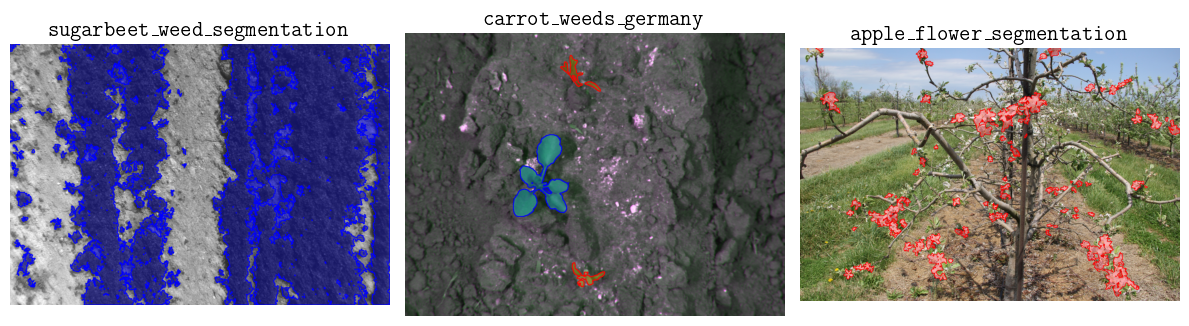}
    \caption{Sample images from a collection of the semantic segmentation datasets used in this research, containing the original image with annotated segmentation masks. The dataset from which each image is sourced from is annotated on top of it.}
    \label{fig:data_diversity_segmentation}
\end{figure}

\begin{figure}[h]
    \centering
    \includegraphics[width=0.95\columnwidth]{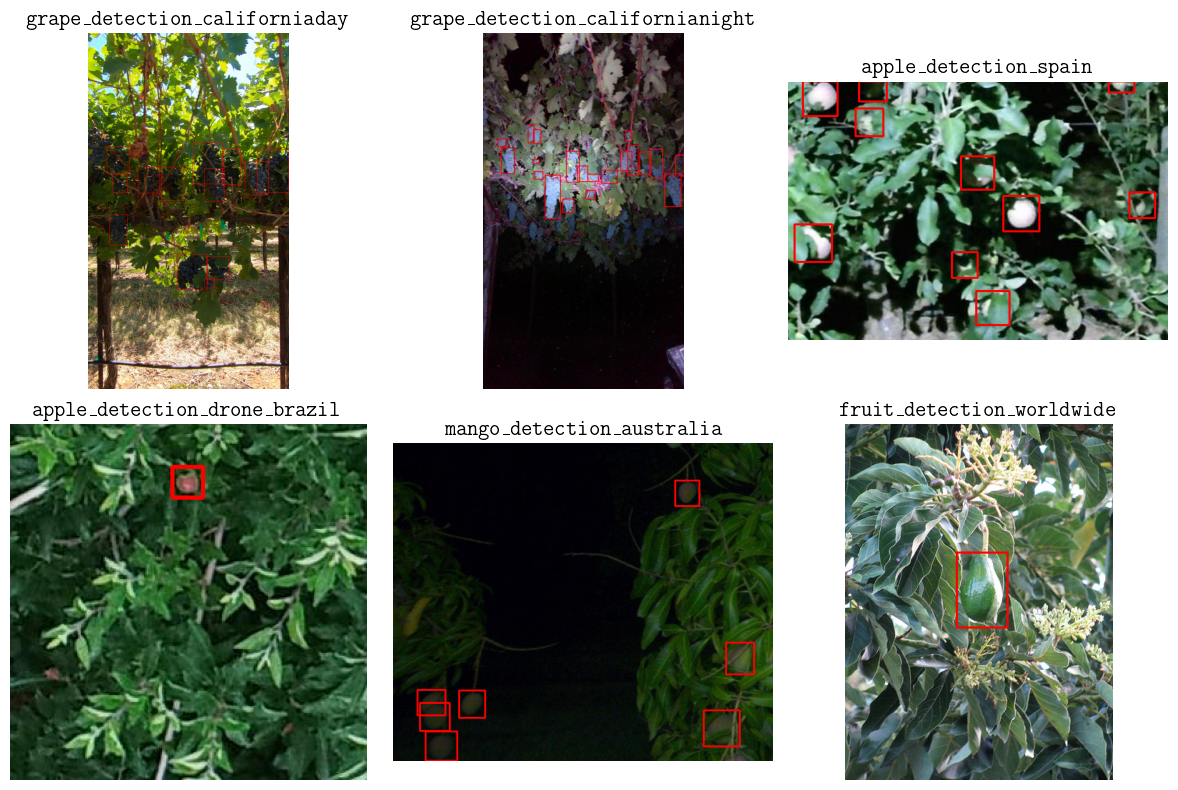}
    \caption{Sample images from a collection of the object detection datasets used in this research, containing the original image with annotated fruit bounding boxes. The dataset from which each image is sourced from is annotated on top of it.}
    \label{fig:data_diversity_object}
\end{figure}

Figure~\ref{fig:data_countries} displays, shaded in green, the countries from which data in AgML's collection of public data sources was gathered from. This data is sourced from each of the six human-inhabited continents. Furthermore, this data covers a wide variety of conditions, including lightning, camera angle, and amount of obfuscation. We illustrate this diversity in two examples. Figure~\ref{fig:data_diversity_object} contains an image from each of the object detection datasets used in our research. Each of the datasets contains a varying set of conditions, ranging from environmental differences, such as day versus night, to camera angle, including ground-level versus drone captures, to the type of fruit (the images shown include apple, grape, mango, and cantaloupe). Similarly, Figure~\ref{fig:data_diversity_segmentation} displays examples of imagery from three semantic segmentation datasets used in our research. We note the variation of agricultural tasks within these datasets, with one dedicated to solely weed segmentation, one being a combination of plant and weed segmentation, and another being flower segmentation. Furthermore, the difference in camera angle and environmental conditions are also present in these datasets.

\subsection{Models and Methods}\label{sec:models_and_methods}

We select our model architectures with two primary focuses: state-of-the-art performance, and ease of access using existing deep learning frameworks. This approach allows us to construct standard pretrained models with both high-scoring benchmarks as well as ease of reproducibility and application to future tasks.
We assess state-of-the-art performance using traditional performance benchmarks, namely ImageNet~\citep{imagenet} for image classification, CityScapes~\citep{cityscapes} for semantic segmentation, and COCO test-dev~\citep{coco} for object detection. Similarly, we define the best-case scenario for ease of access to be the scenario where a model is contained entirely within the PyTorch\citep{pytorch} library family, to reduce external dependencies. We now elaborate on our training procedure for each category of deep learning tasks.

\subsubsection{Image Classification}

In selecting an image classification model, we restrict our search to those available in the \texttt{torchvision} library, in order to facilitate our goal of ease-of-use and reduction of external dependences. The EfficientNet family of models has the highest performance out of all models in the library. So, we select \textbf{EfficientNetB4} as our model for image classification, considering its high ImageNet top-1 accuracy of 82.9\% with a similar number of parameters to traditional state-of-the-art models like ResNet50. We select categorical cross-entropy loss as our criterion, defined as

\begin{equation*}
    L_{CE} = \sum_{i = 1}^n y_i\cdot\log \hat{y}
\end{equation*}

\noindent
where $n$ is the number of classes, $y_i$ is the ground truth, and $\hat{y}$ is the prediction. We evaluate model performance using categorical accuracy.

\subsubsection{Semantic Segmentation}\label{sec:semantic_segmentation_methods}

We again restrict our search to the \texttt{torchvision} library, and in turn we select the \textbf{DeepLabV3} model with a ResNet50 backbone, as described in the original implementation (and as is available in the library). We select a loss function based on the number of classes in the dataset being trained upon. For binary segmentation tasks, involving only a single class, we use binary cross-entropy loss with logits, defined as

\begin{equation*}
    L_{BCE} = -(y\log \hat{y} + (1 - y)\log(1 - \hat{y}))
\end{equation*}

over each pixel, where $y$ represents the ground truth value, and $\hat{y}$ represents the predicted value. For multi-class segmentation tasks, we use dice loss~\citep{dice_loss}:

\begin{equation*}
    L_{D} = 1 - \dfrac{2y\bar{y} + 1}{y + \bar{y} + 1}
\end{equation*}

where $y$ again represents the ground truth and $\hat{y}$ represents the predicted value. We evaluate model performance using the mean Intersection-over-Union (mIoU) metric:

\begin{equation*}
    \mathrm{mIoU} = \dfrac{1}{n + 1} \sum_{i = 0}^n \dfrac{p_{ii}}{\sum_{j = 0}^n p_{ij} + \sum_{j = 0}^n (p_{ji} - p_{ii})}
\end{equation*}

where $n$ is the number of classes and $p_{ij}$ is the number of pixels of class $i$ predicted to belong to class $j$.

\subsubsection{Object Detection}\label{sec:object_detection_methods}

Here, we expand our reach beyond the \texttt{torchvision} library, due to the reduced amount of object detection models available in it. We select the EfficientDet family of models --keeping in line with our prior selection of EfficientNet for image classification -- choosing the \textbf{EfficientDetD4} model, due to its high performance of 49.3\% mAP on the COCO evaluation dataset, with a comparable number of parameters to traditional state-of-the-art models like YOLOv3. We use the open source \texttt{effdet} library\footnote{See: \url{https://github.com/rwightman/efficientdet-pytorch}} as our implementation of the EfficientDetD4 model. The loss function used is focal loss~\citep{focal_oss} with $\alpha = 0.25$ and $\gamma = 1.5$, as described in the original implementation, and given by:

\begin{equation*}
    L_{F} = -(1 - p_t)^\gamma \log(p_t)
\end{equation*}

where $p_t = p$ if $y = 1$, otherwise $p_t = 1 - p$ (and $p$ is the estimated probability for the class $y$). We evaluate model performance using the mean average precision metric, at an IOU threshold of 0.5 (which we denote as \texttt{mAP@0.5}).

\subsection{Model Training Parameters}

For each of the three deep learning tasks, we develop training pipelines which are adaptable to each of the different agricultural datasets. These training pipelines are used both for our benchmark and pretrained model development as well as for our experiments, proving their versatility in adapting to a number of different agricultural deep learning tasks. We summarize the main training parameters for each task in Table~\ref{table:training_details}. Data preprocessing is minimal; our only step involves resizing images to the image size in Table~\ref{table:training_details} and model-specific preprocessing: namely normalization for image classification and semantic segmentation and bounding box format conversion for object detection. In addition, for semantic segmentation and object detection, we decay the initial learning rate by $\gamma = 0.75$ every $8$ epochs. All models are trained using the PyTorch Lightning library~\citep{pytorch_lightning}, on an NVIDIA Titan RTX GPU.

\setlength{\tabcolsep}{4pt}
\begin{table}[!h]
\begin{center}
\caption{Summary of core training parameters for each task.}
\label{table:training_details}
\begin{tabular}{l c l r r}
\hline\noalign{\smallskip}
Task & Image Size & Optimizer & Learning Rate & Epochs \\
\noalign{\smallskip}
\hline
\noalign{\smallskip}

Image Classification & $224$ & Adam & 0.001 & 100\\
Semantic Segmentation & $512$ & NAdam~\citep{nadam} & 0.005 & 50\\
Object Detection & $512$ & AdamW~\citep{adamw} & 0.0002 & 100\\

\hline
\end{tabular}
\end{center}
\end{table}
\setlength{\tabcolsep}{1.4pt}

\subsection{Agricultural Model Weights for Object Detection}

Object detection models are traditionally trained using domain-generic weights: either random initialization or a common benchmark like the COCO dataset. For highly specified environments like agriculture, however, datasets are often of a much smaller magnitude. In many cases, as explored in~\cite{survey_of_public_datasets}, datasets may only consist of a few hundred images, common in the case of object detection where data annotation takes significant time. In turn, using domain-specific agricultural weights as a starting point for deep learning models can significantly boost data efficiency, leading to models attaining high performance with less data and less training time. 

To assess the performance of agricultural pretrained weights, we finetune an EfficientDetD4 model on the \texttt{fruit\textunderscore detection\textunderscore worldwide} dataset, and evaluate a set of five distinct pretrained model approaches: \textit{COCO}, \textit{GRAPE}, \textit{GRAPENIGHT}, \textit{APPLEDRONE} and \textit{NONE}. The parameters of these different models are summarized in Table~\ref{table:pretrained_model_datasets}. \textit{NONE} and \textit{COCO} serve as our baselines for traditional pretrained weight approaches, while \textit{GRAPE}, \textit{GRAPENIGHT}, and \textit{APPLEDRONE} represent agricultural pretrained models trained for different types of environments -- \textit{GRAPE} for daytime, \textit{GRAPENIGHT} for nighttime, and \textit{APPLEDRONE} for drone as opposed to ground imagery. 

\setlength{\tabcolsep}{4pt}
\begin{table}[!h]
\begin{center}
\caption{Summary of pretrained model approaches for single-class detection.}
\label{table:pretrained_model_datasets}
\begin{tabular}{l p{0.4\columnwidth}}
\hline\noalign{\smallskip}
Model Name & Datasets Pretrained On\\
\noalign{\smallskip}
\hline
\noalign{\smallskip}

\textit{NONE} & None\\
\textit{COCO} & COCO~\citep{coco} \\
\textit{GRAPE} & \texttt{grape\textunderscore detection\textunderscore californiaday} \\
\textit{GRAPENIGHT} & \texttt{grape\textunderscore california\textunderscore night} \\
\textit{APPLEDRONE} & \texttt{apple\textunderscore detection\textunderscore drone\textunderscore brazil} \\

\hline
\end{tabular}
\end{center}
\end{table}
\setlength{\tabcolsep}{1.4pt}

We then use these pretrained models for two experiments involving the \texttt{fruit\textunderscore detection\textunderscore worldwide} dataset. We first finetune a model on each of its 7 classes (specifically, 7 one-class models), and then one on the entire 7-class dataset (one 7-class model) -- enabling an assessment of the results of pretrained weights on not just fruit localization, but also classification. Our experimental method and pipeline is carried from Section~\ref{sec:object_detection_methods}, though we restrict our evaluation to the first 50 epochs to obtain a better picture of data efficiency.

\subsection{Agricultural Backbone Weights for Semantic Segmentation}

In contrast to the standard available pretrained weights for object detection, tasks like semantic segmentation often do not have even a domain-general set of pretrained weights available for training -- especially due to the fact that a network with a certain number of classes cannot properly utilize the pretrained weights from one with a different number of classes, unlike in object detection. In such cases, pretrained weights are only used for the \textit{backbone}, a feature extraction model which is used as the input stem into a semantic segmentation network. For instance, in our DeepLabV3 model, the backbone was ResNet50, a model otherwise used for image classification tasks. Traditionally, such classifiers are trained on the ImageNet dataset~\citep{imagenet}. However, as we intend to demonstrate, using agricultural domain-specific weights for even the backbone can performance gain for the larger network as a whole.

\setlength{\tabcolsep}{4pt}
\begin{table}[!h]
\begin{center}
\caption{Summary of pretrained backbones for semantic segmentation. Note that in contrast to object detection, the number of classes of the image classification models are not relevant, as they are being used as feature extractors.}
\label{table:pretrained_model_datasets_segmentation}
\begin{tabular}{l p{0.4\columnwidth}}
\hline\noalign{\smallskip}
Model Name & Datasets Pretrained On \\
\noalign{\smallskip}
\hline
\noalign{\smallskip}

\textit{NONE} & None \\
\textit{IMAGENET} & ImageNet~\citep{imagenet}\\
\textit{VILLAGE} & \texttt{plant\textunderscore village\textunderscore classification} \\
\textit{COUNTING} & \texttt{leaf\textunderscore counting\textunderscore denmark} \\

\hline
\end{tabular}
\end{center}
\end{table}
\setlength{\tabcolsep}{1.4pt}

Image classification datasets are largely different from semantic segmentation datasets -- while most of our semantic segmentation datasets in Table~\ref{table:dataset_listing_with_benchmarks} are for locating fruits or weeds, many our image classification datasets are for \textit{distinguishing} between leaves and weeds. For a thorough examination of the viability of agricultural weights, we select three different agricultural image classification datasets to use as pretrained models, summarized in Table~\ref{table:pretrained_model_datasets_segmentation}. \textit{VILLAGE} is our largest image classification dataset, with over 55,000 images for species and disease classification, while \textit{COUNTING} is a much smaller dataset focused on a largely different application -- leaf counting. We assess these four pretrained backbones on three different datasets: \texttt{apple\textunderscore flower\textunderscore segmentation}, \texttt{apple\textunderscore segmentation\textunderscore minnesota}, and \texttt{rice\textunderscore seedling\textunderscore segmentation}. Our experimental method and pipeline is the same as in Section~\ref{sec:semantic_segmentation_methods}, though we restrict our evaluation to only the first 20 epochs of training.

\subsection{Effectiveness of Augmentations for Generating Diverse Data}

A standard procedure for improving the performance, and in particular, generalizability of deep learning models, is to apply visual and spatial augmentations to the input data, with the intended goal of increasing the diversity of the input data. Standard spatial augmentations include rotation, distortion (such as affine transforms), and cropping, while standard visual augmentations range from saturation and brightness contrast to Gaussian noise addition. For agricultural deep learning models, generalizability is a crucial goal, as environmental conditions vary significantly in different scenarios.

\begin{figure}[!h]
    \centering
    \includegraphics[width=\columnwidth]{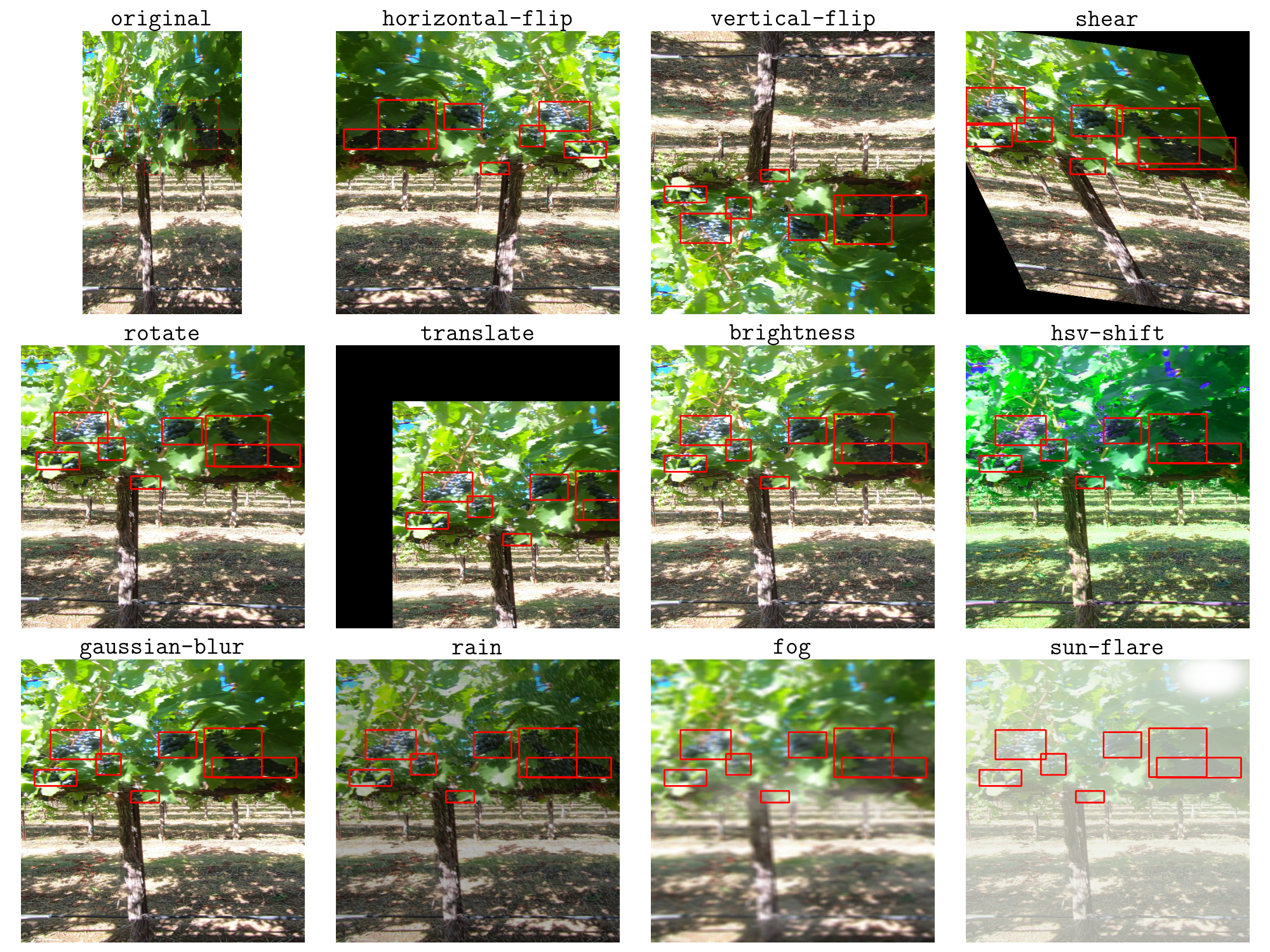}
    \caption{An example image from \texttt{grape\textunderscore detection\textunderscore californiaday} with each of the augmentations used in this research applied to it.}
    \label{fig:augmentation_study}
\end{figure}

In turn, we conduct a robust analysis of various augmentations across a number of datasets in order to collect insight into the most viable augmentations for improving generalizability for agriculture. Performance is assessed of a collection of different augmentations on a set of different datasets across our three different tasks -- image classification, semantic segmentation, and object detection -- to provide a greater insight into each augmentation's effectiveness. Our analysis only involves the usage of a single augmentation per trial, without combinations, in order to provide an insight as to the capability of each augmentation independently, relevant to our goal of assessing augmentations in a agricultural context. Additionally, we collect final metric scores early in training, namely at only 20 epochs. As noted from our prior observations, most models begin approaching the relative peak of their performance at this threshold. So, a better analysis of the effectiveness of augmentations is done by analyzing their performance improvements earlier in training. We use the albumentations~\citep{albumentations} library for our augmentations, noting its seamless integration into our existing PyTorch training pipeline. The augmentations we use, alongside model performance corresponding to the datasets we select, are summarized in Table~\ref{table:augmentation_experiment} to eliminate redundancy. Furthermore, we provide a visual example of each of the augmentations we use in Figure~\ref{fig:augmentation_study}.

\subsection{Effects of Annotation Quality on Learning}

Data annotation is a heavily time-consuming task within deep learning. In the field of agriculture, this is an especially prevalent issue, due to the present nuance and even simply the quantity of objects such as fruits which need to be annotated in fruit detection tasks. Oftentimes, the amount of effort needed to annotate a certain dataset can result in that dataset being generated with imperfect and even obviously inaccurate annotations, potentially leading to them being unusable for deep learning tasks. However, with the already present scarcity of agricultural data, any loss of existing data has larger impacts on the open-source data community. Thus, we now assess how the quality of annotations on agricultural data affects models' overall performance. In particular, we aim to observe whether models are capable of achieving high performance without perfect annotations, potentially leading to less strict annotation requirements for deep learning models in the future.

We use three fruit detection datasets: \texttt{grape\textunderscore detection\textunderscore californiaday}, \texttt{apple\textunderscore detection\textunderscore drone\textunderscore brazil}, and \texttt{fruit\textunderscore detection\textunderscore worldwide}. For each dataset, we set aside a common test set. For the remaining data, we create a set of datasets with reduced quality annotations, removing a certain percentage of bounding boxes for each image. This procedure is conducted for retention of only 30\% of annotations, to 90\% of annotations, with 10\% increments. Using our object detection pipeline from Section~\ref{sec:object_detection_methods}, we train for 25 epochs on each dataset and record their \texttt{mAP@0.5} on the aforementioned test sets.

\section{Results and Discussion}

\subsection{Standard Benchmarks for Agricultural Datasets}

The performance of each of our task-based models on the datasets we have collected is summarized in Table~\ref{table:dataset_listing_with_benchmarks}. Our standard pipelines enable our models to achieve comparable performance with existing benchmarks on our collected datasets, in certain cases even exceeding them. As mentioned in Section~\ref{sec:models_and_methods}, these datasets, along with pretrained models which achieved these benchmarks, are available through the open-source framework AgML, enabling further research on developing even more efficient data and model pipelines.

\subsection{Performance of Agricultural Pretrained Weights for Object Detection}

\begin{figure}[!h]
    \centering
    \includegraphics[width=\columnwidth]{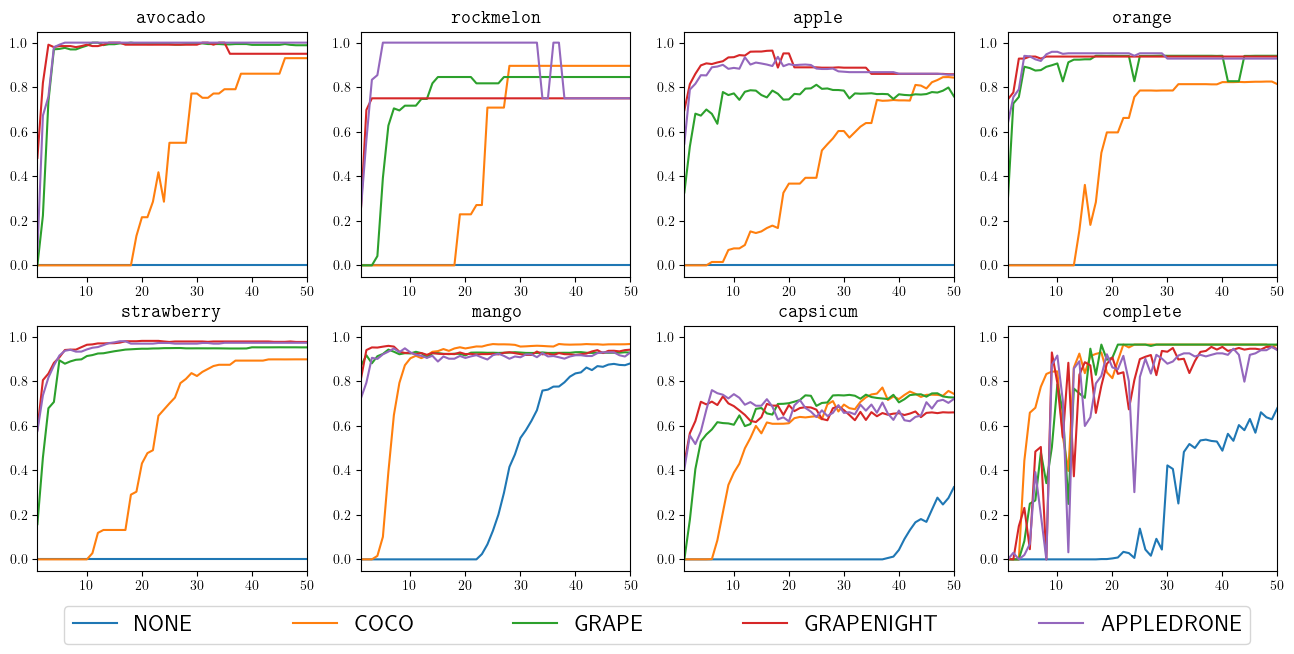}
    \caption{Comparison of \texttt{mAP@0.5} values on seven different individual fruits from the \texttt{fruit\textunderscore detection\textunderscore worldwide} dataset, alongside the complete dataset (listed as \texttt{complete}, the bottom right plot).}
    \label{fig:detection_pretrained_graph}
\end{figure}

A summary of the results of our experiments using agricultural pretrained weights for object detection are shown in Figure~\ref{fig:detection_pretrained_graph}, with each of the fruits representing the 7 one-class models, and \texttt{complete} representing the one 7-class model. For all of the 7-class models, the pretrained agricultural models significantly outperform the \textit{COCO} and \textit{NONE} baselines. We observe that the agricultural models tended to plateau at their maximum mean average precision before 10 epochs, for most fruits -- some, like strawberry, avocado, and mango, see this plateau as early as 5 epochs. On the other hand, \textit{COCO}, the standard baseline, usually takes between 25 to 40 epochs to reach its own maximum value, while the \textit{NONE} model fails to even break zero mean average precision for five out of seven fruits, and only reaches a comparable score to \textit{COCO} for one. For the 7-class model, we see that the agricultural pretrained models follow a similar trajectory to \textit{COCO}, taking around or over 30 epochs to plateau at their maximum mean average precision.

A number of observations can be drawn from these results. Pretrained weights are clearly a better approach when training object detection models, as they converge significantly faster than random weights, saving time and resources in training. The poor performance of the \textit{NONE} model demonstrates the value of transferring prior knowledge when training on new tasks. Furthermore, for agricultural domain-specific detection tasks, using agricultural pretrained weights can provide, at minimum, comparable performance to \textit{COCO}, or other standard weights, but at best, even better performance and much faster convergence. Our results are consistent with previous observations for other tasks, such as~\cite{plant_disease_transfer_learning}, where agricultural pretrained weights were found to increase performance. Our results are further consistent with previous observations specifically for object detection such as in~\cite{transfer_learning_for_pears}, where using pretrained fruit detection weights from similar environments resulted in a significant performance increase. A key insight from our, work, however, is the fact that agricultural environment played little role in performance improvement -- all three models, despite their distinct environments, achieved similar performance on a dataset in yet another noticeably different environment. Furthermore, each agricultural pretrained model followed a similar performance \textit{trajectory}, corroborating our observations that using agricultural pretrained weights of any form result in not only higher performance, but faster convergence. In turn, this approach, which only requires a slight modification to existing pipelines, can significantly improve training results for future agricultural deep learning work.

\subsection{Performance of Agricultural Backbone Weights for Semantic Segmentation}

\begin{figure}[!h]
    \centering
    \includegraphics[width=\columnwidth]{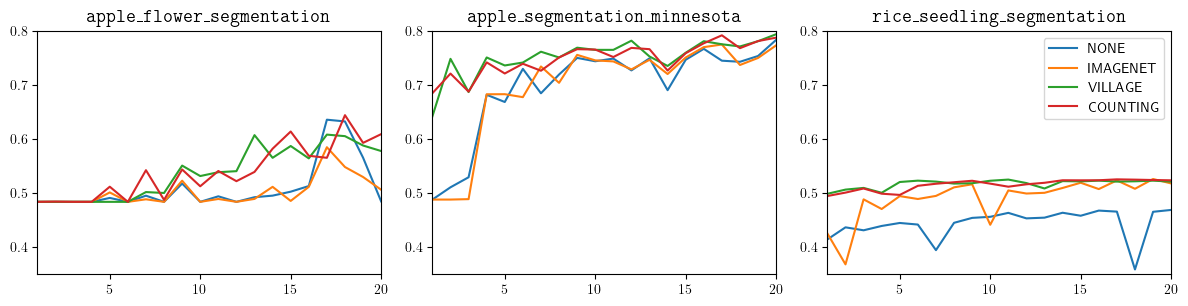}
    \caption{Comparison of \texttt{mIOU} values for four different pretrained backbones on three different evaluation datasets. The $y$-axis is on a consistent scale, although each dataset has its models converge to a different range of values.}
    \label{fig:sem_seg_backbone_graph}
\end{figure}

Similar to the previous section, we summarize the results of our semantic segmentation models in Figure~\ref{fig:sem_seg_backbone_graph}. We find that our models with agricultural pretrained backbones, \textit{VILLAGE} and \textit{COUNTING}, tend to outperform models with the existing \textit{NONE} and \textit{IMAGENET} backbones to a measureable degree. For the datasets \texttt{apple\textunderscore segmentation\textunderscore minnesota} and \texttt{rice\textunderscore seedling\textunderscore segmentation}, our agricultural pretrained models reach their maximum \texttt{mIOU} almost within the first couple of epochs, while the other two models either take between 10-15 epochs to reach a similar degree of performance, or do not reach it at all. For \texttt{apple\textunderscore flower\textunderscore segmentation}, our models reach similar levels of performance as \textit{COCO}, but maintain a more consistent trajectory of performance increase, while \textit{COCO} and \textit{NONE} see a spike in performance followed by a quick drop, suggesting relative inconsistency.

Semantic segmentation is a unique task due to the required level of data annotation required for it -- each pixel in an image must be assigned to a certain class. This is reflected in the reduced size of segmentation datasets, an example of which can be seen in the image counts of Table~\ref{table:dataset_listing_with_benchmarks}. As a result, there is potentially not enough available data to generate a large-scale pretrained network for semantic segmentation, especially when noting the potential difference in classes between a pretrained network and an applied, task-specific network -- resulting in loss of knowledge in weights which cannot be transferred. This is noted in works like~\cite{deep_agronomic_segmentation} and~\cite{grape_segmentation_varieties}, where standard domain-general weights are used in lieu of their agricultural equivalents. However, image classification datasets are in relative surplus in comparison to semantic segmentation, and most segmentation models use classification models as their feature extractors, representing the encoder module for a typical encoder-decoder architecture. For most tasks, a standard approach is to freeze weights for this feature extraction module, the encoder, and only update parameters for the segmentation head, the decoder. This is inspired by the fact that the feature extraction model obtains the most relevant information regarding the location of the objects being segmented -- however, for agricultural networks, this approach seems largely counter-intuitive, as the feature extractor is usually trained on domain-general images and this unable to recognize the most relevant features for plants. Using agricultural pretrained weights, on the other hand, can improve the localization performance of the feature extractor, in turn improving the network's performance as a whole. Our results show that using agricultural weights for feature extractors serves as a potential workaround for the lack of a standard semantic segmentation pretrained model, still enabling improved performance and reduced training time.

\subsection{Augmentation Effectiveness}

We summarize the performance of each of the augmentations used in this work, on six different datasets ($A, B, C$ object detection, and $D, E, F$ semantic segmentation), in Table~\ref{table:augmentation_experiment}.
When using a single augmentation, as in our experiments, observe that spatial augmentations -- namely augmentations (1) through (5) as denoted in the table, when used independently -- generally outperform visual augmentations -- (6) through (11) -- for all of the datasets used in this experiment. The augmentations which provide the largest performance increase are all spatial, while the visual augmentations, in many cases, fail to even provide \textit{any} performance increase whatsoever. For our subsequent analysis, we refer to each dataset using its key in the table (\textit{A} refers to \texttt{grape\textunderscore detection\textunderscore californiaday}, and so on).

\setlength{\tabcolsep}{4pt}
\begin{table}[!h]
\begin{center}
\caption{Summary of performance of different augmentations on the aforementioned semantic segmentation and object detection datasets. Bold indicates the highest performance for a certain dataset.}
\label{table:augmentation_experiment}
\begin{tabular}{l r r r r r r}
\hline\noalign{\smallskip}
Augmentation & \textit{A} & \textit{B} & \textit{C} & \textit{D} & \textit{E} & \textit{F}  \\

\noalign{\smallskip}
\hline
\noalign{\smallskip}

 & \multicolumn{3}{c}{\texttt{mAP@0.5}} & \multicolumn{3}{c}{\texttt{mIOU}} \\
\hline
\noalign{\smallskip}

\texttt{original} (0) & 64.08\% & 53.62\% & 75.11\% & \textbf{69.78\%} & 78.52\% & 51.04\% \\
\texttt{horizontal-flip} (1) & 62.76\% & 58.32\% & 74.46\% & 66.97\% & 79.19\% & \textbf{51.35\%} \\
\texttt{vertical-flip} (2) & \textbf{68.45\%} & 56.12\% & 74.48\% & 62.48\% & 75.35\% & 49.19\% \\
\texttt{shear} (3) & 63.31\% & 58.28\% & 80.96\% & 65.61\% & 77.54\% & 51.04\% \\
\texttt{rotate} (4) & 60.77\% & \textbf{61.06\%} & 84.42\% & 64.51\% & 73.96\% & 49.75\% \\
\texttt{translate} (5) & 65.53\% & 53.05\% & \textbf{86.60\%} & 66.26\% & \textbf{79.22\%} & 48.57\% \\
\texttt{brightness} (6) & 59.01\% & 52.69\% & 78.80\% & 66.45\% & 77.72\% & 44.50\% \\
\texttt{hsv-shift} (7) & 56.51\% & 49.32\% & 82.39\% & 51.34\% & 73.12\% & 42.95\% \\
\texttt{gaussian-blur} (8) & 65.17\% & 57.09\% & 84.60\% & 69.25\% & 78.61\% & 50.96\% \\
\texttt{rain} (9) & 60.44\% & 42.25\% & 81.20\% & 69.49\% & 73.09\% & 50.11\% \\
\texttt{fog} (10) & 65.35\% & 48.66\% & 78.81\% & 62.73\% & 70.81\% & 44.31\% \\
\texttt{sun-flare} (11) & 65.92\% & 55.87\% & 80.72\% & 49.86\% & 74.70\% & 43.63\% \\

\noalign{\smallskip}
\hline
\noalign{\smallskip}
\hline
\noalign{\smallskip}

\multicolumn{2}{l}{\multirow{6}{*}{Dataset Legend}} & \multicolumn{5}{r}{$^{A}$\texttt{grape\textunderscore detection\textunderscore californiaday}}\\
&& \multicolumn{5}{r}{$^{B}$\texttt{grape\textunderscore detection\textunderscore californianight}} \\
&& \multicolumn{5}{r}{$^{C}$\texttt{apple\textunderscore detection\textunderscore drone\textunderscore brazil}} \\
&& \multicolumn{5}{r}{$^{D}$\texttt{apple\textunderscore segmentation\textunderscore minnesota}} \\
&& \multicolumn{5}{r}{$^{E}$\texttt{apple\textunderscore flower\textunderscore segmentation}} \\
&& \multicolumn{5}{r}{$^{F}$\texttt{rice\textunderscore seedling\textunderscore segmentation}} \\
\hline

\end{tabular}
\end{center}
\end{table}
\setlength{\tabcolsep}{1.4pt}

Our next key observation regards the effect of different augmentations on different environments and conditions. For instance, datasets \textit{A} and \textit{B} consist of the same fruit in the same environment, simply in day conditions versus night conditions. While certain visual augmentations, such as rain and fog, provide a performance boost for \textit{A}, they actually significantly reduce performance for \textit{B}, indicating that certain factors, such as rain and fog, are less valuable information in nighttime environments as opposed to daytime environments. Another example of the effects of different augmentations, this time for spatial augmentations with reference to camera positioning, is found when observing the results for $C$. While $A$ and $B$, the other object detection datasets, consist of ground cameras capturing images of plants, $C$ consists of aerial imagery captured by a drone. In turn, augmentations like rotation, translation provide a better opportunity for the model to generalize to different aerial camera positions, as opposed to less distortive augmentations like horizontal and vertical flips. For semantic segmentation tasks, we observe a less obvious boost in performance from augmentations. In fact, for $D$, no augmentations provide the best result, while for $E$ and $F$ the performance boost is within one percent of no augmentations. This potentially stems from the different conditions of semantic segmentation tasks as opposed to object detection tasks -- semantic segmentation tasks, especially for fruit or leaf segmentation, like $D$ and $E$, can come down to individual pixels, and such precision may likely be distorted by spatial augmentations. In all tasks, while visual augmentations appear to provide less of an obvious benefit as opposed to spatial augmentations, our results still provide some insight into their potential viability. In particular, augmentations which affect the hue of an image, such as HSV shift and brightness or contrast shifts, tend to actually decrease performance, as they result in a model adapting to those input colors. On the contrary, Gaussian blurring provides a consistent, though sometimes minute, performance increase relative to the input. This suggests that a reduction of features, as done by adding a slight blur to the input, can in fact increase performance by potentially having the model learn more general features in the input environment.

A major benefit of augmentations is their ability to generate data potentially modified for a large variety of conditions, enabling greater generalizability for agricultural models. Existing state-of-the-art work done for domain transfer often involves the usage of generative adversarial networks (GANs), such as~\cite{enlisting_gans_generalizability}, who developed a GAN for transferring sample imagery between day and night domains, and~\cite{domain_adaptation_cyclegan}, who used a CycleGAN network to edit the fruits present in imagery while maintaining the environmental conditions. While augmentations may not necessarily be able to provide an entire domain transfer as in the prior methods, they can still provide a generalization of conditions -- for instance, rain and fog augmentations, as used in our research, can generalize data to a broader range of common conditions across the world. Other augmentations, like sun flare, can reduce the impact of edge cases where images are obscured. 
We note that certain augmentations expand beyond a single data points, potentially involving multiple images being transformed together. This is the case in common augmentations such as mosaic and patch, as explored in their applications for agricultural robotics in~\cite{augmentation_for_crop_robotics}. While we do not include these augmentations in our research, as they involve larger modifications to our pipelines, they do present potential further performance improvements for agricultural models.  If used properly based on the input environment, augmentations provide an efficient way to produce a larger set of environments for agricultural data, improving generalizability of models.

\subsection{Annotation Quality}

\begin{figure}[!h]
    \centering
    \includegraphics[width=\columnwidth]{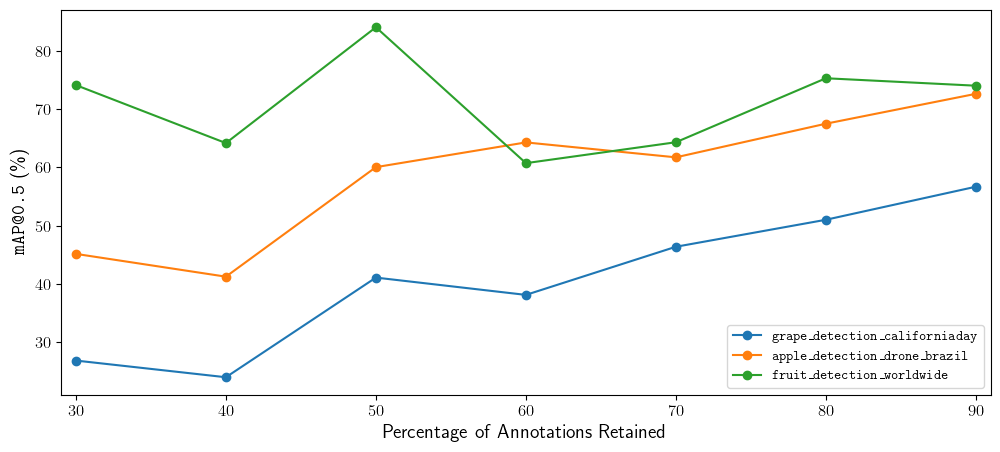}
    \caption{Performance of models on three different datasets dependent on the percentage of bounding box annotations retained per image.}
    \label{fig:annotation_quality_graph}
\end{figure}

\begin{figure}[!h]
    \centering
    \includegraphics[width=\columnwidth]{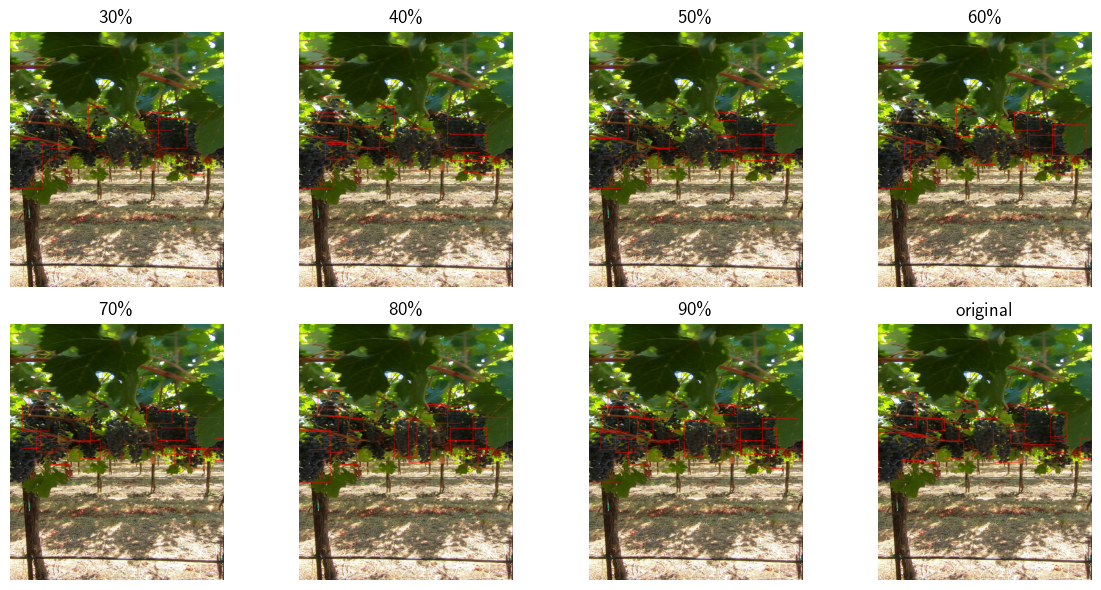}
    \caption{Sample predictions by models trained on different levels of annotation quality. The percentages above each image reflect the amount of bounding boxes retained for the data used in training that specific model.}
    \label{fig:annotation_quality_grape_sample}
\end{figure}

The performance of each of our annotation quality models are recorded in Figure~\ref{fig:annotation_quality_graph}, which displays the performance of models relative to the quality of annotations on the datasets they are trained on. For the datasets \texttt{grape\textunderscore detection\textunderscore californiaday} and \texttt{apple\textunderscore detection\textunderscore drone\textunderscore brazil}, we find a consistent increase in performance as the quality of annotations increase. Notably, we do not observe a one-to-one correspondence between quality of annotations and level of performance: e.g., 30\% of annotations actually corresponds to over 50\% of peak performance, suggesting that models trained on lower quality annotations are, at least to some extent, able to localize similar objects to a slightly higher than expected degree. We provide a sample of predicted bounding boxes for each \texttt{grape\textunderscore detection\textunderscore californiaday} iteration in Figure~\ref{fig:annotation_quality_grape_sample}. This observance, however, does not hold for \texttt{fruit\textunderscore detection\textunderscore worldwide}, which has a distinction of being a multi-class dataset in contrast to the prior two. Surprisingly, for this dataset, we find that our models achieve maximum performance with only 50\% of annotations. This example, in turn, provides an insight into the hazardous potential of using lower-quality annotations. At 50\% of annotations, our model may have only learned general features, thus allowing it to still perform much better on other sample images. However, higher percentages of still low-quality annotations may result in a model learning to distinguish between different instances of the same fruit by extremely nuanced features, overfitting on the samples on which it is trained.

A number of methods have been proposed in recent work to try and automate the data annotation process, which is one of the major bottlenecks for agricultural deep learing. Some approaches, such as~\cite{robotic_data_labeling}, use robotic systems to capture data from a wider range of angles and automatically annotate bounding boxes knowing the location of plants in controlled environments. Other approaches~\citep{semi_supervised_smart_ag,pseudo_label_generation} involve semi-supervised learning, taking advantage of highly precise data to train high-performance models which in turn improve predictions on unlabeled images. Such approaches show potential for not only making new, high-quality agricultural datasets, but even potentially improving upon existing datasets, annotating missed fruits or other objects which may have been missed out on. Nevertheless, our work demonstrates that fruit detection models can still obtain relatively high performance with some fruits unannotated. In turn, potential new approaches may involve using lower-quality datasets in coordination with better datasets, expanding the pool of available data for agricultulture and further boosting model performance.

\section{Conclusion}

In our work, we have developed a novel set of standardized and centralized agricultural datasets, alongside benchmarks and pretrained models using state-of-the-art models. Our custom pipelines achieve comparable performance with existing benchmarks using no extensive data or architectural modifications, making them widely applicable to a variety of agricultural deep learning tasks. We have also assessed a number of existing methods for improving model performance domain-specific to agriculture, including using agricultural pretrained model weights and image augmentations. Furthermore, we have even explored the viability of traditionally overlooked lower-quality data, potentially expanding the data pool for agriculture. Our results demonstrate that slight training modifications can significantly boost model performance and result in shorter convergence time. We have open-sourced our standardized and centralized versions of the datasets used in our work, alongside our pretrained models and benchmarks, to guide easier adoption of our described methods.

\section{Acknowledgements}

This project was partly supported by the USDA AI Institute
for Next Generation Food Systems (AIFS), USDA award
number 2020-67021-32855.

\clearpage

\bibliographystyle{mla}
\bibliography{egbib}
\end{document}